\documentclass[letterpaper]{article} % DO NOT CHANGE THIS
\usepackage{aaai25}  % DO NOT CHANGE THIS
\usepackage{times}  % DO NOT CHANGE THIS
\usepackage{helvet}  % DO NOT CHANGE THIS
\usepackage{courier}  % DO NOT CHANGE THIS
\usepackage[hyphens]{url}  % DO NOT CHANGE THIS
\usepackage{graphicx} % DO NOT CHANGE THIS
\usepackage{natbib}  % DO NOT CHANGE THIS AND DO NOT ADD ANY OPTIONS TO IT
\usepackage{caption} % DO NOT CHANGE THIS AND DO NOT ADD ANY OPTIONS TO IT
\usepackage{algorithm}
\usepackage{algorithmic}
\usepackage{amsmath}
\usepackage{multirow}
\usepackage{booktabs}
\usepackage{bm}
\usepackage{amssymb}

\usepackage{newfloat}
\usepackage{listings}
\DeclareCaptionStyle{ruled}{labelfont=normalfont,labelsep=colon,strut=off} % DO NOT CHANGE THIS
\floatstyle{ruled}
\newfloat{listing}{tb}{lst}{}
\floatname{listing}{Listing}
\title{Local Conditional Controlling for Text-to-Image Diffusion Models}
\author{Yibo Zhao$^{1,2}$\quad Liang Peng$^{2}$\quad Yang Yang$^{1,2}$\quad Zekai Luo$^{1,2}$\quad Hengjia Li$^{1,2}$\quad Yao Chen$^{2,3}$\\ Zheng Yang$^{2}$\quad Xiaofei He$^{1,2}$ \quad Wei Zhao$^{4}$\quad Qinglin Lu$^5$\quad Boxi Wu$^{3}$\quad Wei Liu$^5$\\ 
}

\affiliations{
    \textsuperscript{\rm 1}State Key Lab of CAD\&CG, Zhejiang University, \quad \textsuperscript{\rm 2}Fabu Inc.\\
    \textsuperscript{\rm 3}The School of Software Technology, Zhejiang University, \quad 
    \textsuperscript{\rm 4}Xidian University, \quad
    \textsuperscript{\rm 5}Tencent Inc.\\ 

}

\usepackage{bibentry}
\begin{document}

%\maketitle

\twocolumn[{
\renewcommand\twocolumn[1][]{#1}
\maketitle
\begin{center}
    \centering
    \vspace*{-1.0cm}
    \includegraphics[width=\textwidth]{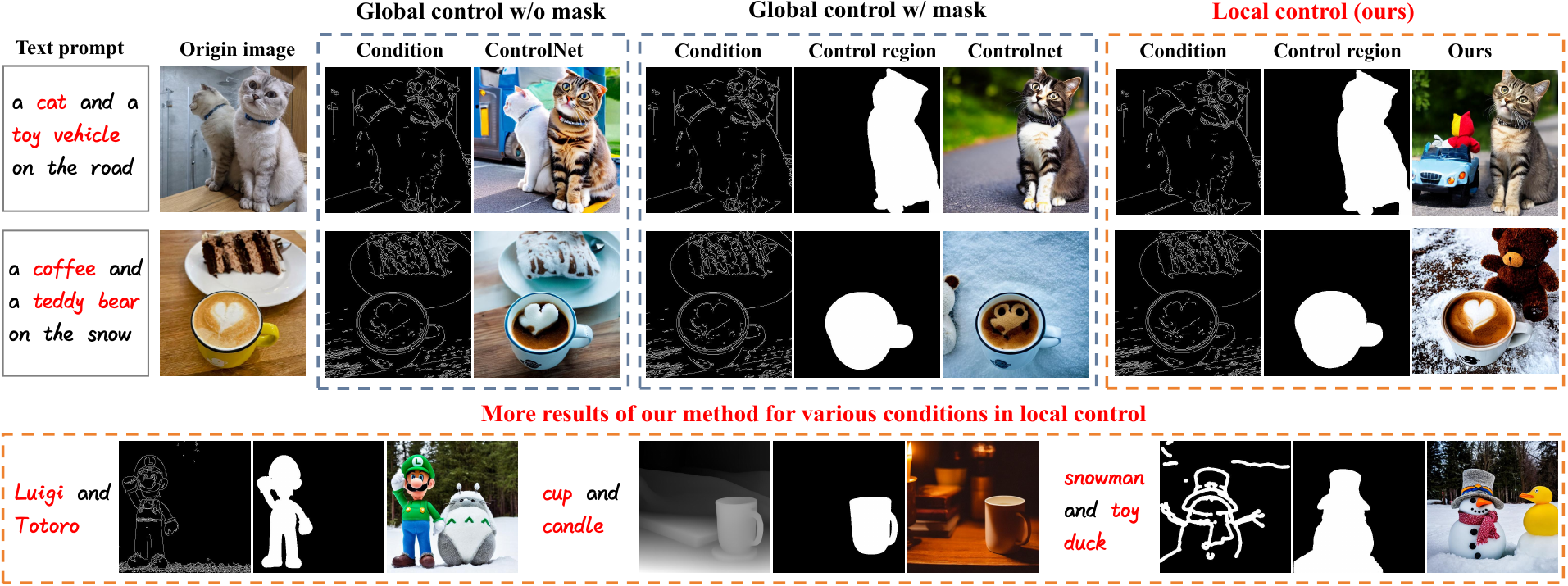}
    \vspace*{-.6cm}
    \captionof{figure}{
    Previous global control mechanism mainly synthesizes images similar to structure conditions, but has difficulty generating results aligned with text prompts.
    Even adding control mask could only produce concepts closest to the local condition.
    Therefore, we explore local control, which leverages text prompts, image conditions, and user-defined regions for local control as inputs. 
    Our proposed method successfully generate images that are faithful to both the prompts and local control conditions.
    }
\label{fig:head}
\end{center}
% \vspace*{-.6cm}
}]
\begin{abstract}
Diffusion models have exhibited impressive prowess in the text-to-image task. Recent methods add image-level structure controls, \textit{e.g.}, edge and depth maps, to manipulate the generation process together with text prompts to obtain desired images.
This controlling process is globally operated on the entire image, which limits the flexibility of control regions.
In this paper, we explore a novel and practical task setting: \textbf{local control}.
It focuses on controlling specific local region according to user-defined image conditions, while the remaining regions are only conditioned by the original text prompt.
However, it is non-trivial to achieve local conditional controlling. 
The naive manner of directly adding local conditions may lead to the local control dominance problem, which forces the model to focus on the controlled region and neglect object generation in other regions.
To mitigate this problem, we propose Regional Discriminate Loss to update the noised latents, aiming at enhanced object generation in non-control regions.
Furthermore, the proposed Focused Token Response suppresses weaker attention scores which lack the strongest response to enhance object distinction and reduce duplication.
Lastly, we adopt Feature Mask Constraint to reduce quality degradation in images caused by information differences across the local control region.
All proposed strategies are operated at the inference stage.
Extensive experiments demonstrate that our method can synthesize high-quality images aligned with the text prompt under local control conditions.
\end{abstract}

\section{Introduction}
\label{sec:intro}

Large-scale text-to-image diffusion models \cite{saharia2022photorealistic,rombach2022high,ramesh2021zero} have showed remarkable capabilities in generating images conditioned on specified text prompts. However, text descriptions frequently fall short in sufficiently conveying detailed control \cite{wang2022pretraining} over the generation images. Recent methods \cite{zhang2023adding,mou2023t2i} have introduced the incorporation of image-level spatial controls, such as edges, depth, and segmentation, into the text-to-image generation process. These methods operate across the entire image, ensuring that the final synthesized image retains the complete global structural information of the original. Such advancements have attracted considerable attention from both the academic and industrial sectors, as it facilitates a more controllable image synthesis process.

\begin{figure*}[t]
    \centering
    \includegraphics[width=0.75\linewidth]{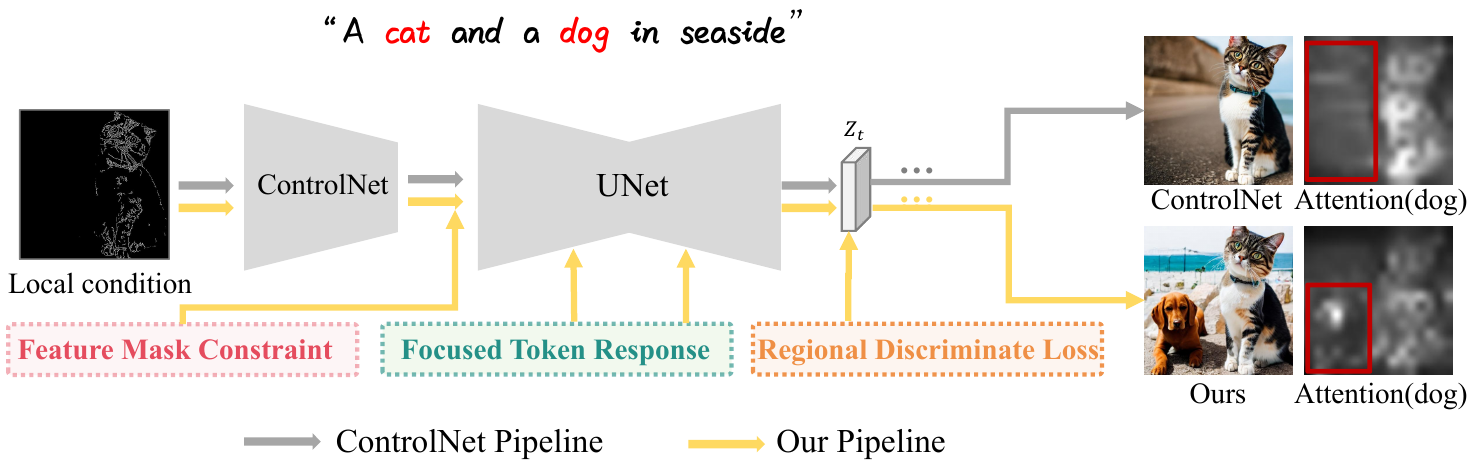}
    \vspace{-7pt}
    \caption{
    % {\bf Local control dominance.} 
    In the ControlNet pipeline, local conditions dominate the image generation. The attention map for ``dog'' exhibits a higher response in local areas while a lower response in non-local regions where the dog should be generated.
    Meanwhile, we showcase the result of our method applied to local control , where both ``cat''  and ``dog'' are successfully generated.
    }
    \label{fig:cross_attention}
 \vspace{-16pt}
\end{figure*}

Unfortunately, the conventional approach of enforcing global control at the entire-image level lacks the fine-grained image manipulation capability that users often require. 
Considering the example in Figure \ref{fig:head}, we might want to exclusively apply the cat canny condition, while allowing the generation in other regions fully leverages the capabilities of the text-to-image model.
This naturally leads to a question: Can we confine our control to specific local region using image conditions?
In this paper, we specifically address this practical scenario and introduce the concept of local control as a novel and pragmatic paradigm. 
It focuses on controlling the generation of specific regions of an image according to user-defined local control conditions, while the rest of the image is conditioned by the text prompt.
As illustrated in Figure \ref{fig:head}, only the canny condition of the cat is employed for controlling, while the remainder of the image is generated solely based on the provided text prompt. 
This novel manner empowers users with the ability to exert flexible control over the image generation process.

An intuitive resolution for local control lies in directly adding local image conditions.
However, this naive method leads to a catastrophic problem.
The prompt concept that is most related to the local control condition dominates the generation process, while other prompt concepts are ignored.
Consequently, the generated image cannot align with the input prompt.
We call this problem \textbf{local control dominance}.
We show the cases in the Figure \ref{fig:cross_attention}. 
Delving into the underlying mechanism, we find that the attention map for the dog shows high response within the local control regions, which impedes the generation of multiple objects, as seen here with the dog's generation being hindered. 
It indicates that local control conditions force the model to focus on the controlled region and overlook other regions.

To address this issue, we introduce an object regeneration strategy. Specifically, we propose the Regional Discriminate loss to update the latents.
It identifies the ignored objects in the cross attention map, enhancing them on the latent by adaptively operating attention scores under local and non-local regions.
This manner guides the model to generate the most related objects to the image condition in the local region, meanwhile producing other desired objects in non-control regions for the text prompt.
Additionally, we suppress attention scores for patches on the cross attention map which lack the strongest response in the token dimension. This technology alleviates token overlapping contribution to enhance object distinction and reduce duplication.
Moreover, we use local region mask to constrain the feature outputs of ControlNet. This technology reduces the degradation of synthesized image quality caused by differences in control information between non-local and local control regions.
All proposed strategies are operated at the inference stage.
Overall, the proposed training-free approach achieves high-quality local control for both the text prompt and image conditions, without the need for any training or additional text-image paired datasets. 
Extensive experiments demonstrate the effectiveness of our method.

Our contributions can be summarized as follows:
\begin{itemize}
\item
We introduce a new task setting: local control, allowing for fine-grained controlling with image conditions.
It aims to control specific regions within the image as desired by the user, while the remaining regions are conditioned solely by the original text prompt.
\item
We pose the main challenge in local control, and propose a training-free method that includes three techniques to resolve it.
The proposed method ensures the preservation of local control while maintaining the generative capability of the diffusion models.
\item
Extensive experiments demonstrate that our method can synthesize high-quality images, precisely aligning both local image conditions and text prompts. 
\end{itemize}

\section{Related Work}
\label{sec:relat}

\textbf{Text-to-Image Models.} Early methods \cite{xu2018attngan,zhang2017stackgan} for text-to-image synthesis are limited in generating low-resolution images in specific domains.In recent years, extensive data volume LAION\cite{schuhmann2022laion} and abundant model capacity have led to significant progress in the field of text-to-image synthesis. The diffusion model \cite{dhariwal2021diffusion,ho2020denoising} has achieved amazing results in text-to-image generation tasks and has gradually become a representative method in this field. Such as Glide\cite{nichol2021glide}, Imagen \cite{saharia2022photorealistic}, Stable Diffusion \cite{rombach2022high}, and DALL-E \cite{ramesh2021zero}. They leverage the denoising process, gradual reduction of noise to high-quality images through iterative optimization. In addition, the cross-attention mechanism incorporates content from text prompts into the generated images through CLIP\cite{radford2021learning} or large language models\cite{raffel2020exploring}. 

\noindent\textbf{Compositional Generation.} 
Compositional Generation involves constructing or combining images semantically based on a set of given elements or attributes.
\cite{yang2023paint,avrahami2022blended,lu2023tf} utilize diffusion models to tackle the problems of fusion inconsistency, enhancing the efficiency of image composition. 
Recent works focus on mutil-concept generation, Prompt2Prompt \cite{hertz2022prompt} and Plug-and-Play \cite{Plug-and-Play_2023_CVPR} explore the relationship between attention map and image representation.
StructureDiffusion \cite{feng2022training}, Attend-and-Excite \cite{chefer2023attend} and Multi-Concept T2I-Zero \cite{tunanyan2023multi} refine the cross-attention map to strengthen all of the subjects to encourage the model to generate all concepts described in the text prompts. 
Gligen \cite{li2023gligen} achieves grounded text2img generation with text prompt and bounding box condition inputs.
Boxdiff \cite{xie2023boxdiff} propose a training-free method to control objects and contexts in the synthesized images. 
Unlike these methods, our approach aims to preserve the structural information of local regions during image generation, which facilitates finer-grained generation.
 
\noindent\textbf{Controllable Generation.} 
To facilitate controllable image generation, some works aim to provide reliable structural guidance for the synthesis results. PITI \cite{wang2022pretraining} proposes to enhance structural guidance by reducing the gap between features derived from conditions. \cite{voynov2023sketch} suggests using the similarity gradient between the target sketch and intermediate model features to impose constraints. ControlNet \cite{zhang2023adding} learns condition-specific network to enable guide generation for the pre-trained diffusion model. T2i-Adapter\cite{mou2023t2i} introduces a lightweight adapter to combine the internal knowledge in T2I models with control signals. 
Uni-ControlNet\cite{uni} proposes a condition injection strategy to learn an adapter for various controls, focusing on combined generation of multiple structure conditions.
% freecontrol
Recently, training-free approaches \cite{masa,patashnik2023localizing,epstein2023diffusion,freecontrol} are proposed for controllable T2I diffusion.
Although these existing structure-controllable image generation methods have achieved good control at the image-level, they lack the ability to provide localized control for generating images that align with both prompts and control conditions.

\section{Method}
\label{sec:int}
\begin{figure*}[!htt]
    \centering
    \includegraphics[width=\linewidth]{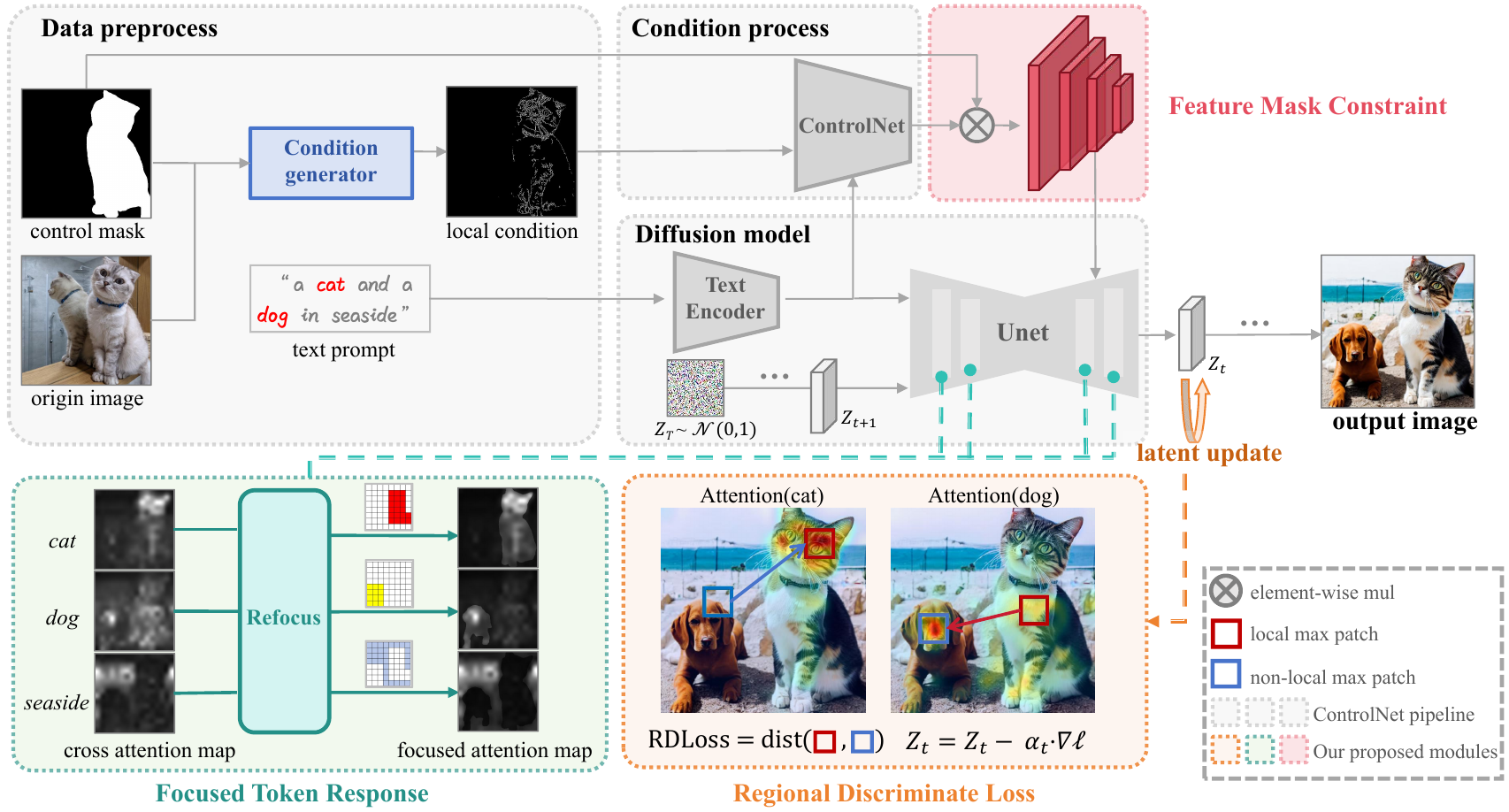}
    \vspace{-20pt}
    \caption{
    Overview of our method. 
    Given the text condition and local control condition, a latent variable \( z_T \) is passed into the denoising network, \emph{i.e.,} UNet. 
    At each denoising step, we apply \textbf{Feature Mask Constraint} (see Section \ref{sec:Feature}) to the ControlNet branch output features. 
    Cross-attention maps generated from UNet are refined by \textbf{Focused Token Response} (see Section \ref{sec:Focused}), which suppresses weaker interfering responses to enhance object distinction.
    Furthermore, we use \textbf{Regional Discriminate Loss} (see Section \ref{sec:Attention}) to update the latent \( z_t \), thereby identifying and regenerating ignored objects in the cross-attention map. } 
    \label{fig:pipeline}
 \vspace{-15pt}
\end{figure*}

In this section, we first introduce the preliminary knowledge required for local control.
Secondly, we present Control Concept Matching which identifies objects for local control(see Section \ref{sec:Control}).
We then detail our proposed training-free method that synthesizes images using the local control condition and the text prompt as inputs.
% This method builds upon prior research \cite{hertz2022prompt} which has shown that the prominently responsive regions within the cross-attention map are crucial in determining the spatial positioning of generated objects.
Our method is illustrated in Figure \ref{fig:pipeline}. 
Specifically, our method consists of three components: 
1) \textbf{Regional Discriminate Loss} (see Section \ref{sec:Attention}), which updates the latents to facilitate the generation of ignored objects; 
2) \textbf{Focused Token Response} (see Section \ref{sec:Focused}), which suppresses non-maximum attention scores to enhance object distinction; 
3) \textbf{Feature Mask Constraint} (see Section \ref{sec:Feature}), which mitigates image quality degradation caused by discrepancies in control information.

\subsection{Preliminaries}
\label{sec:pre}
\textbf{Stable Diffusion.} 
We apply our method on Stable Diffusion \cite{rombach2022high}, which efficiently operates on the latent space. Specifically, given an input image $I$, the encoder maps it to a latent $z_0$. 
Diffusion algorithms progressively add noise $\epsilon$ to the $z_0$ and produce a noisy latent $z_t$, where $t$ represents the number of times noise is added.
To achieve conditional generation, Stable Diffusion utilizes cross-attention between text embeddings \(\mathbf{c}\) converted by CLIP \cite{clip} and the UNet's intermediate features \(\varphi(\mathbf{x}_t)\). The cross-attention map \(\mathbf{A}\) can be defined as:
\begin{equation}
    \mathbf{A} = \text{Softmax}(\mathbf{Q}\mathbf{K}^{\top} / \sqrt{d}),
\end{equation}
where \(\mathbf{Q} = \mathbf{W}_Q \varphi(\mathbf{x}_t)\) and \(\mathbf{K} = \mathbf{W}_K \mathbf{c}\) are the query and key matrices, respectively. Finally, $\epsilon_\theta$ can be trained by a MSE loss,  which predicts the noise added to the latent $\bm{z}_t$:
\begin{equation}
    \mathcal{L} = \mathbb{E}_{\bm{z}_0, \bm{t}, \bm{c},\epsilon \sim \mathcal{N}(0, 1) }\Big[ \Vert \epsilon - \epsilon_\theta(\bm{z}_{t}, \bm{t}, \bm{c}) \Vert_{2}^{2}\Big]
\end{equation}

\noindent\textbf{ControlNet.} 
Relying solely on text prompts often fails to generate images that align with the user's desires. ControlNet\cite{zhang2023adding} integrate global-level spatial controls which utilizes the encoding layers of the diffusion model.
During training, an additional spatial control condition \(\bm{c}_\text{f}\) is required, and the training loss is:
\begin{equation}
    \mathcal{L}_c = \mathbb{E}_{\bm{z}_0, \bm{t}, \bm{c}, \bm{c}_\text{f}, \epsilon \sim \mathcal{N}(0, 1) }\Big[ \Vert \epsilon - \epsilon_\theta(\bm{z}_{t}, \bm{t}, \bm{c}, \bm{c}_\text{f})) \Vert_{2}^{2}\Big].
\end{equation}
Therefore, ControlNet can utilize both text and image structure as conditions to control image generation.

In the context of local control, given a text prompt, the tokenizer produces a set of tokens \(\mathcal{P} = \{\mathbf{p}_i\}\), and a set of corresponding spatial cross-attention maps \(\mathcal{A}^{t} = \{\mathbf{A}^{t}_i\}\) in UNet can be accordingly obtained.
The most straightforward method of enabling local control is to directly input \(\bm{c}_\text{f}\) and $\mathcal{P}$ into the existing control methods. However, this approach leads to the issue of local control dominance (Figure \ref{fig:cross_attention}).

\subsection{Control Concept Matching}
\label{sec:Control}
Our initial objective in the image synthesis process with local control is to identify the most suitable object for generation within the control region at timestep t.
The resulting object token indices are denoted as \(\text{C}^{t}_\text{control}\). 
In our method, the sum of attention scores within the local control region is employed as the criterion.
At denoising steps \( t > \beta T \), we identify the object with the highest sum attention score within the local control region as the \(\text{C}^{t}_\text{control}\). 
$\beta$ is a hyperparameter that acts on the total timesteps $T$.
Given the impact of the denoising process's initial steps on the spatial structure of the synthesized image, we focus on stabilizing the selection of \(\text{C}^{t}_\text{control}\). 
Thus, we employ the \(Count_{max}(\cdot)\) function to determine the most frequently occurring \(\text{C}^{t}_\text{control}\) in the early steps of the denoising process.

\begin{equation}
    \text{C}^{t}_{\text{control}} = 
    \begin{cases} 
    {\arg\max}_i \Big(\sum (\mathbf{M} \cdot \mathbf{A}^{t}_{i}) \Big),& t > \beta T, \\
   Count_{max}(\{ \text{C}^{t^{*}}_{\text{control}} \mid t^{*} > \beta T \}),& t \leq \beta T 
    \end{cases}
\end{equation}
where $\mathbf{M}$ denotes a binary spatial mask, which is generated from the local control region.
It aims to selectively local control elements in the cross-attention maps. 
This approach ensures that the chosen object is optimally aligned with the user-defined local control conditions.

\subsection{Regional Discriminate Loss}{\label{sec:Attention}}

Due to the impact of local control dominance, most objects often respond to the local control regions in cross-attention maps. An effective strategy to counter this effect involves directing the attention responses of objects in non-control regions away from the local control region. To achieve this, we introduce the object regeneration strategy, which is specifically designed to enhance the difference in response between the inside and outside of the local control region. 

Attend-and-excite\cite{chefer2023attend} indicates that the maximum value regions in the cross-attention map are key to defining the positioning of generated objects. Therefore, we choose the maximum value in the attention map to design a loss, named Regional Discriminate Loss (RDLoss). We adopt the max distance between the maximum values inside and outside the local region as the loss, so that each attention map can separate the objects in the local and non-local regions based on whether they are \(\text{C}^{t}_\text{control}\) adaptive.

\begin{equation}
\begin{aligned}
    \text{RDLoss}_{s_{i}} &= \mathbb{I}\left( \max(G(\mathbf{A}_{s_{i}}) \cdot \mathbf{M}) \right. \\
    &\quad \left. - \max(G(\mathbf{A}_{s_{i}}) \cdot (1 - \mathbf{M})) \right),
    \label{eq:l1}
\end{aligned}
\end{equation}
we select a set of object token indices \(\mathcal{S} = \{\mathbf{s}_i\}\) from the collection \(\mathcal{P}\), where \(\mathcal{P}\) represents the entire set of prompt tokens. \(G(\cdot)\) represents a Gaussian smoothing operator. And the indicator function \(\mathbb{I}\) is defined as follows:
\begin{equation}
\mathbb{I}(s_{i} = \text{C}^{t}_\text{control}) =
\begin{cases}
-1, & \text{if } s_{i} = \text{C}^{t}_\text{control}, \\
1, & \text{otherwise}.
\end{cases}
\end{equation}
RDLoss uses the difference between the maximum attention scores inside and outside the local region to represent distance.
In the attention maps for tokens not requiring local control conditions, we enhance the generation in non-local regions. Conversely, for tokens that require local control, we enhance generation in local regions.
After defining the loss, we can update the latent in the following way: 

\begin{equation}
    z_t = z_t - \alpha_{t} \nabla l,
\end{equation}
$l$ is determined by \( \max(RDLoss_{si}) \) in eq \ref{eq:l1} and is used to update the noised latent $z_t$ at the current time step \( t \), where \( \alpha_{t} \) is a scalar defining the step size of the gradient update. $l$ updates noised latents by increasing the attention scores that exhibit the maximum difference between patchs inside and outside the local control region.
This approach directs the model to generate objects closely related to the control condition in the local region, while also creating objects in the non-control regions as given text prompt.

\subsection{Focused Token Response}{\label{sec:Focused}}
After updating the noised latent, while the image content closely aligns with the text prompt, non-local objects still tend to be generated repeatedly, or near the local control region.
This challenge is often linked to the overlapping contribution regions of corresponding token embeddings.
Inspired by\cite{tunanyan2023multi}, we utilize cross attention maps to mitigate weak response regions. Specifically, we reduce the attention scores for patchs in the attention map that do not demonstrate the strongest response in the token dimension. This technology encourages the reduction of responses from weaker tokens, thereby enforce the model to selectively respond to the most relevant tokens.

\begin{equation}
\label{eq:l2}
\mathbf{A}^{t} = O(\arg\max(\mathbf{A}^{t}),\gamma) \cdot \mathbf{A}^{t},
\end{equation}
\( \arg\max\) identifies the indices of the highest responses across token dimension, while \( O \) represents a suppression operation that applies a scaling factor \( \gamma \) to the non-max attention scores. 
This technique minimizes the influence of irrelevant pixels in the cross-attention map, enhances specific token contributions, reduces overlap and duplication.
Collaborating with RDLoss, it aids the model in generating across both local and non-local regions.

\subsection{Feature Mask Constraint}{\label{sec:Feature}}
During the denoising process, there is a notable difference in the amount of information between inside and outside of the local control region. The local control carries only localized control information, leaving the rest of the regions with blank control information. As a consequence, the latent distribution resulting from the injection of local control conditions into the UNet differs from that obtained under global control conditions, potentially affecting the image quality.

To address this issue, we apply feature mask constraint to the output of ControlNet. 
This design specifically directs the UNet's focus to the local control region when adding control. It thus minimizes the disruptive effect of irrelevant blank control information in non-control region. In practical terms, we extract the local feature maps from the output of ControlNet block. Subsequently, these local condition features are added in the UNet. Assume \(\mathcal{F}(\cdot; \theta)\) represents a neural block within the UNet architecture, and \(x_t\) is the input feature map of the block during the \(t\)-th denoising step.
Separately, \(\mathcal{G}(\cdot; \cdot)\) represents a neural block in ControlNet, which takes the spatial conditional input feature \(c_{f}\) and \(x_t\).
The fused feature output from each block is denoted as \(y_t\):
\begin{equation}
y_t = \mathcal{F}(x_t;\theta) + \mathbf{M} \cdot \mathcal{G}(x_t;c_{f}).
\end{equation}
This technique is applied in the computation process of the RDLoss.
After applying the feature mask constraint, the output from the ControlNet is selectively integrated into the UNet. The local control condition is achieved through a multiplication operation involving the feature mask \(\mathbf{M}\) and the conditioned output from the ControlNet block \(\mathcal{G}(x_t;c_{f})\). This combination is then added to the original feature map \(\mathcal{F}(x_t;\theta)\) from the UNet. 
This method improves image quality by explicitly preventing the UNet from focusing on irrelevant control information.

\begin{figure}[t]
\centering
\includegraphics[clip,width=\columnwidth]{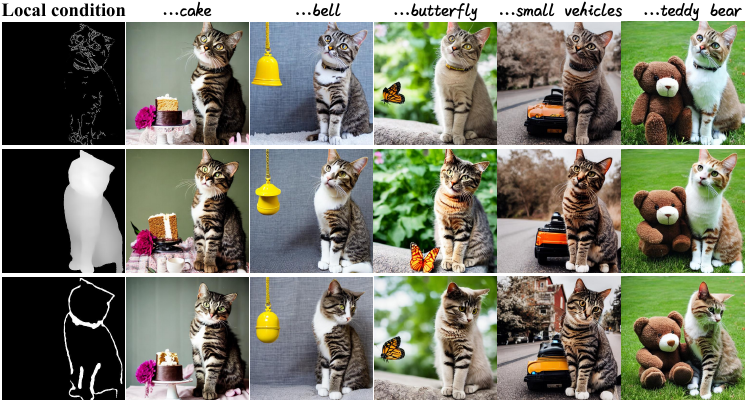}
    \vspace{-17pt}
    \caption{
    Visual results of our method under the local control setting.
    The proposed method can be employed for flexibly controlling the local regions with image conditions. 
    } 
    \label{fig:datu}
  \vspace{-15pt}
\end{figure}

\begin{figure*}[!htt]
 \vspace{-20pt}
    \centering
    \includegraphics[width=\linewidth]{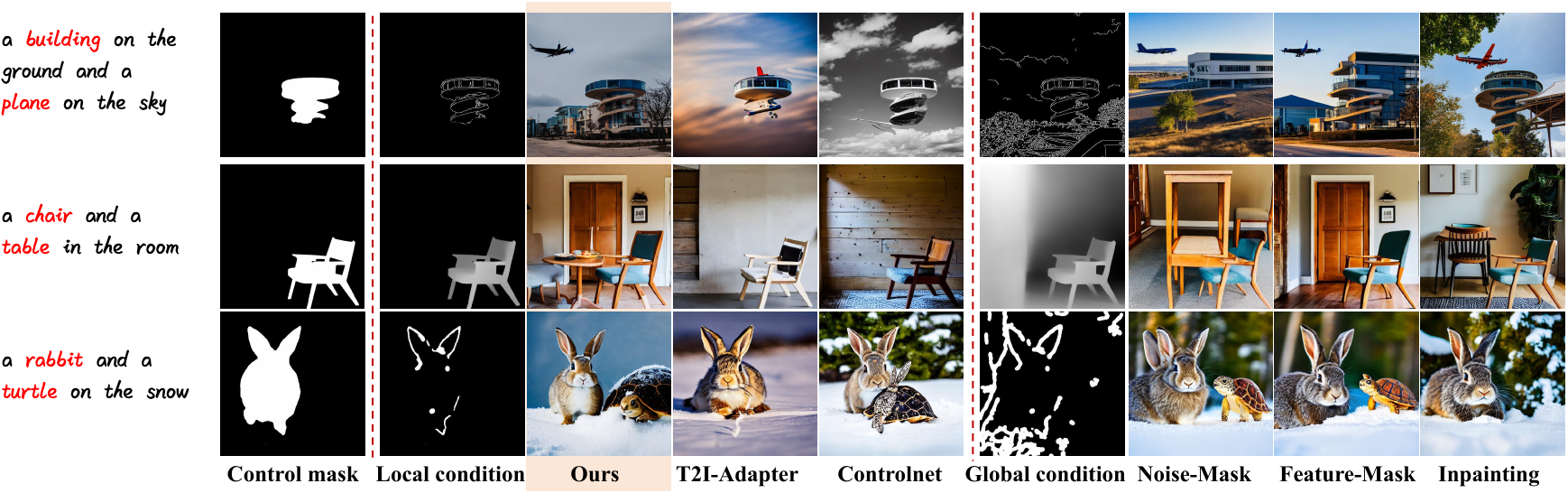}
    \vspace{-18pt}
    \caption{
    Comparisons with several baselines under varying conditions.
    Our method, along with T2I-Adapter and ControlNet, uses local conditions as input. Conversely, Noise-Mask, Feature-Mask, and Inpainting use global conditions as inputs.
    All these methods use the same control mask. Compared with these baselines, our method can synthesize high-quality images in local control, especially in terms of text alignment.
    } 
    \label{fig:baselines}
\end{figure*}

\begin{table*}[!ht]
     \vspace{-4pt}
    \renewcommand{\arraystretch}{1.2}
    \centering
    \resizebox{\linewidth}{!}
    {\begin{tabular}{@{}cc|cccccc@{}}
            \toprule
            \multirow{1}{*}{Dataset}
            & \multirow{1}{*}{metrics}
            & ControlNet & T2I  & Noise-Mask & Feature-Mask & Inpainting &  Ours \\

            \midrule
            \multirow{6}{*}{COCO2017}
            & FID $\downarrow$ 
            & 27.87 $\pm$ 4.22 & 30.32 $\pm$ 3.75 & 22.72 $\pm$ 0.61 & 21.93 $\pm$ 0.64 & 25.72 $\pm$ 1.22 &  \textbf{21.86 $\pm$ 0.48} \\
            
            & CLIP Score $\uparrow$ 
            & 0.293 $\pm$ 0.010 & 0.295 $\pm$ 0.03 & 0.287 $\pm$ 0.003 & 0.290 $\pm$ 0.002 &  0.288 $\pm$ 0.003 & \textbf{0.305 $\pm$ 0.02} \\

            & CLIP T2T $\uparrow$ 
            & 0.782 $\pm$ 0.019 & 0.780 $\pm$ 0.009 & 0.789 $\pm$ 0.006 & 0.779 $\pm$ 0.014  & 0.785 $\pm$ 0.006 & \textbf{0.801 $\pm$ 0.006} \\
            
            & IOU $\uparrow$
            & \textbf{0.55 $\pm$ 0.07} &0.44 $\pm$ 0.06 & 0.41 $\pm$ 0.02 & 0.37 $\pm$ 0.03 &0.53 $\pm$ 0.07 & 0.51 $\pm$ 0.10 \\

            & LPIPS$_{^{\times 10^2}}$ $\downarrow$ 
            & 13.15 $\pm$ 1.19 & 14.00 $\pm$ 1.06 & 14.16 $\pm$ 1.12 & 15.01 $\pm$ 1.33 & 13.25 $\pm$ 1.51 &  \textbf{13.04 $\pm$ 1.36} \\
            \bottomrule

            \hline
            \multirow{6}{*}{Attend-Condition}
            & FID $\downarrow$ 
            & -  & -  & - & -  & -  & -\\
            
            & CLIP Score $\uparrow$ 
            & 0.261 $\pm$ 0.010 & 0.258 $\pm$ 0.004 & 0.279 $\pm$ 0.003 & 0.278 $\pm$ 0.003 & 0.273 $\pm$ 0.03 & \textbf{0.285 $\pm$ 0.02} \\

            & CLIP T2T $\uparrow$ 
            & 0.732 $\pm$ 0.018  & 0.700 $\pm$ 0.015 & 0.782 $\pm$ 0.006 & 0.784 $\pm$ 0.019 & 0.753 $\pm$ 0.009 & \textbf{0.804 $\pm$ 0.002} \\

            & IOU $\uparrow$
            & \textbf{0.74 $\pm$ 0.04} &0.70 $\pm$ 0.02 &0.66 $\pm$ 0.02 & 0.64 $\pm$ 0.04 &0.72 $\pm$ 0.04 & 0.70 $\pm$ 0.05 \\

            & LPIPS$_{^{\times 10^2}}$ $\downarrow$
            & 12.67 $\pm$ 1.03 & 13.31 $\pm$ 1.15 &13.25 $\pm$ 1.21 & 13.30 $\pm$ 1.27 & 12.97 $\pm$ 1.06 & \textbf{12.57 $\pm$ 1.08} \\
            \bottomrule
            
    \end{tabular}}
     \vspace{-5pt}
    \caption{Quantitative comparisons to baselines. We do not provide FID on the Attend-Condition dataset due to the small scale of its original images. Our method can maintain good alignment with the provided spatial condition while achieving the best image quality and text alignment.}
    \label{table:baseline}
     \vspace{-15pt}
\end{table*}

\begin{figure}[!htt]
\centering
\includegraphics[clip,width=\columnwidth]{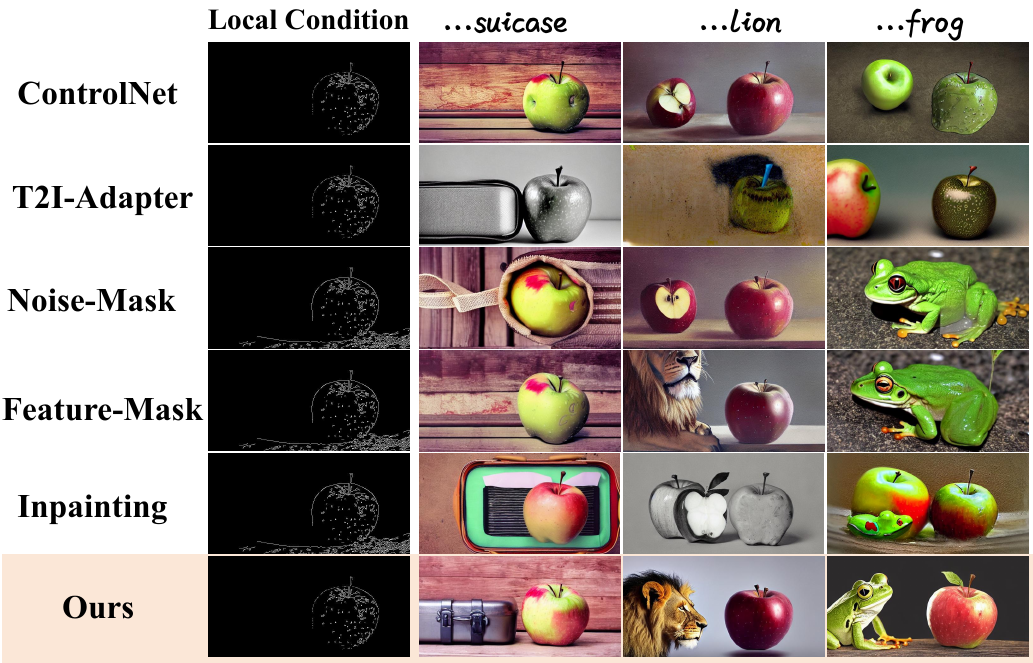}
     \vspace{-20pt}
    \caption{
Comparisons with baselines on Attend-Condition.
    } 
    \label{fig:compare2}
  \vspace{-17pt}
\end{figure}

\section{Experiments}

\subsection{Dataset and Evaluation}
{\label{sec:dataset}}

\textbf{Dataset:}
We utilized the COCO\cite{coco} validation set, which comprises 80 object categories. Each image in this set is associated with multiple captions, we randomly selected one caption per image, resulting in a dataset of 5k generated images. Additionally, we create a dataset inspired by \cite{chefer2023attend}, named \textbf{Attend-Condition}. This dataset encompasses 11 object and 11 animal categories, with text prompts typically structured as `a [object] and a [animal]'. We collect images of each concept from the internet and randomly select those with a larger subject area to serve as representative images for local control.

\noindent\textbf{Evaluation Metrics:}
For assessing image quality, we used the Fréchet Inception Distance (FID)\cite{fid}, while the average CLIP\cite{clips} score between text prompts and corresponding images measured text-image similarity. Additionally, the primary focus of our evaluation is on whether the generated images align with the objects mentioned in the captions. Utilizing the descriptive capabilities of BLIP-2\cite{li2023blip}, we generate captions for the images produced under the local control paradigm. Subsequently, we calculate the CLIP text-text similarity between these generated captions and the original captions, providing a comprehensive measure of content alignment with the text prompt.
To evaluate the alignment with the provided spatial condition, we caculate the LPIPS\cite{lpips} metrics between original and generated images in localized regions. 
To evaluate if local object generation matches specified regions, we use the IOU metric. we adopt the segmentation-based evaluation approach used in ControlNet on the COCO dataset, which uses OneFormer\cite{oneformer} to detect the local segmentations then compute the IOU. Additionally, we use SAM\cite{sam} and DINO\cite{dino} to segment the Attend-Condition experimental results.

\noindent\textbf{Baseline Settings} We conducted a comparative analysis of our method against several baselines.
We select two commonly used structure-controllable models, ControlNet\cite{zhang2023adding} and T2I-Adapter\cite{mou2023t2i}, as naive baselines by inputting the local control condition and text prompts into these models.
Additionally, we design three competitive baselines specifically for the local control task, which do not require extra training and use global conditions as input. Specifically, 1)Noise-Mask: Inspired by \cite{nichol2021glide}, we estimate the noise with adding the condition into ControlNet and the noise from vanilla T2I model, then combine them according to the mask. The combined noise is calculated as:
\begin{equation}
\epsilon_\theta = \epsilon_\theta(\bm{z}_{t}, \bm{t}, \bm{c}, \bm{c}_\text{f}) \cdot \mathbf{M} + \epsilon_\theta(\bm{z}_{t}, \bm{t}, \bm{c}) \cdot (1-\mathbf{M}).
\label{eq:loss}
\end{equation}
2)Feature-Mask: We implement the addition of our proposed “Feature Mask Constraint” component into ControlNet across all denoising steps, which applies local control mask to ControlNet features. 
3)Inpainting: We use ControlNet to generate an image with the spatial condition and text prompt, then use inpainting method\cite{lugmayr2022repaint} to regenerate the parts without spatial condition.

\subsection{Comparison with Baselines}

\begin{figure*}[!ht]
 \vspace{-20pt}
    \centering
    \includegraphics[width=\linewidth]{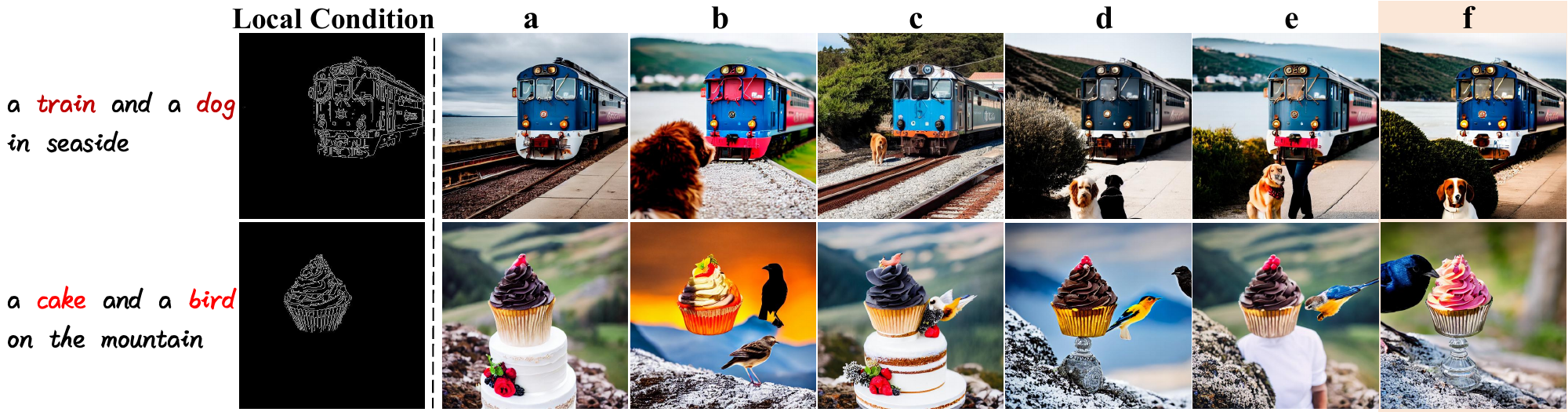}
    \vspace{-18pt}
    \caption{
    Ablation Studies: a) We use ControlNet\cite{zhang2023adding} as the vanilla baseline. b) Only RDLoss is included. c) Only the FMC is included, referred to as the feature-mask baseline. d) Our method without the FTR module. e) Our method without the FMC module. f) Our complete method. 
    } 
    \label{fig:ablation}
    \vspace{-12pt}
\end{figure*}

\noindent\textbf{Quantitative experiments.} We evaluate the baselines and our method under various conditions such as canny, depth, and sketch on the COCO and Attend-Condition datasets. The result includes the following metrics: FID, CLIP-score, CLIP text-text similarity (CLIP T2T), IOU, and LPIPS, which are detailed in Table~\ref{table:baseline}. 
Due to the small number of original images in the Attend-Condition, we do not calculate its FID, as such a small sample size leads to unreliable FID.
In certain cases, the local control region is significantly smaller than the entire condition image, which results in many blank pixels. This leads to the naive methods generating artifacts and grayscale images, which result in high FID scores.
% Figure \ref{fig:baselines} R2C3 and Figure \ref{fig:ablation} R3C3 showcased failure examples.
In evaluations on Attend-Condition dataset specifically designed for multi-object generation, our method significantly outperforms other baselines in both CLIP Score and CLIP T2T.
Additionally, we evaluate the alignment ability of baselines and our method. Noise-Mask and Feature-Mask methods align poorly with the given local spatial conditions. ControlNet achieves the best spatial alignment performance. The pipeline of the inpainting method ensures spatial alignment capabilities similar to that of ControlNet.
Moreover, our method also achieves similar spatial alignment capabilities as ControlNet in local control task. 
Compared to these baselines, our method maintains good alignment with the given spatial conditions, while achieving the best image quality and text alignment.

\noindent\textbf{Qualitative Results.} 
Figure \ref{fig:baselines} illustrates these differences, particularly in scenarios involving multiple objects in prompts. 
In naive method, it can almost only generate objects in text prompt that best match the control region, while neglecting the generation of other objects. The other three baselines designed specifically for local control achieved good text-image alignment and image quality, but have significant drawbacks compared to our visualized results.
Noise-Mask frequently generates untrustworthy pixels near the local region, and sometimes fails to generate mutilple objects.
The results generated by the Feature-Mask approach align poorly with the given spatial conditions.
The inpainting method sometimes fails to distinguish between the generated content within controlled and uncontrolled regions, leading to issues of repetition and omission in the generated images. For instance, an additional chair in the second row, and a turtle is missing in the third row. 
Compared to these baselines, our method generates images that most closely match the text prompts.

\subsection{Ablation Studies}
\noindent\textbf{Main ablations.} 
We utilize the canny condition on the COCO dataset to evaluate the effectiveness of our method's key components. Specifically, we focus on three components: 1) Regional Discriminate Loss(RDLoss), updates the noised latent to assist object generation in non-local regions; 2) Focused Token Response(FTR), designed to refine attention response; and 3) Feature Mask Constraint(FMC), a mechanism to constraint features of the ControlNet output. Only the FMC is included, referred to as the feature-mask baseline. Additionally, since the FTR depends on RDLoss, there is no ablation study for it alone. The images are generated using the same seed. The results are listed in Table~\ref{table:Ablation}. 
It reveals that RDLoss reduces the influence of local control dominance in non-local regions. FTR and FMC can assist RDLoss in improving the quality of generated images.
It can be foreseen that the updates to the latent have somewhat diminished the capability for image alignment, yet the extent of this impact remains within an acceptable threshold. 
Figure \ref{fig:ablation} visualizes the results of the ablation experiments.

\begin{table}[!htt]
  \renewcommand{\arraystretch}{0.5}
  \centering
    \small
  \resizebox{\linewidth}{!}{
  \begin{tabular}
{ccc|c@{\hspace{0.7em}}c@{\hspace{0.7em}}c@{\hspace{0.7em}}c@{\hspace{0.7em}}c@{\hspace{0.7em}}c@{\hspace{0.7em}}}
    \toprule
    RDLoss & FTR & FMC & FID $\downarrow$
            & CLIP Score$\uparrow$
		& CLIP T2T$\uparrow$
            & IOU$\uparrow$
            & LPIPS$_{^{\times 10^2}}$$\downarrow$ \\
    \midrule
     - & -  & - & 23.65 & 0.283 & 0.750 & \textbf{0.62} &  11.96\\
     
    $\surd$ & - & - & 24.48 & 0.299 & 0.786 & 0.59&  11.99\\
    - & - & $\surd$ & 21.83 & 0.292 & 0.791 & 0.40&  13.68\\
    $\surd$ & - & $\surd$ &  21.91 & 0.301 & 0.802 & 0.57 &  11.71\\
    $\surd$ & $\surd$ & - &  22.90 & 0.298 & \textbf{0.806} & 0.59 &  11.79\\
    $\surd$ & $\surd$ & $\surd$ &  \textbf{21.38} & \textbf{0.303} & 0.795 & 0.57 &  \textbf{11.68}\\   

    \bottomrule
  \end{tabular}}
  \vspace{-8pt}
  \caption{Ablation studies of the proposed components under canny condition on the COCO dataset.
  Since the FTR depends on RDLoss, there is no ablation study for it alone. 
  }
  \label{table:Ablation}
  \vspace{-4pt}
\end{table}

\begin{figure}[h]
\centering
\includegraphics[clip,width=\columnwidth]{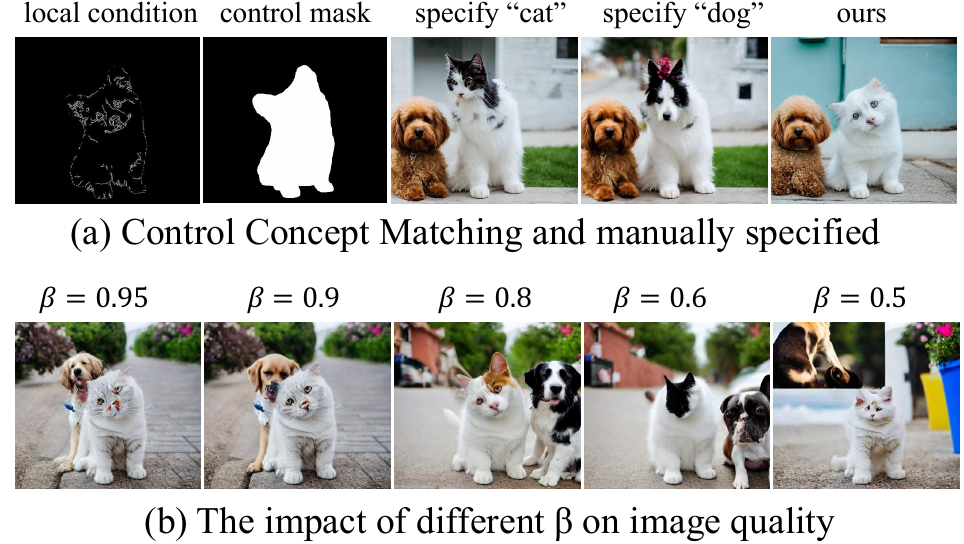}
\vspace{-15pt}
    \caption{ (a) Control Concept Matching adaptively matches the concept, (b) Different \(\beta\) values affect image quality.} 
    \label{fig:beta}
\vspace{-16pt}
\end{figure}
\noindent\textbf{Impact of Control Concept Matching.}
The Figure \ref{fig:beta}(a) shows the necessity  of selecting concepts which match the local control region. 
% The middle images correspond to manually specified \(\text{C}^{t}_\text{control}\) of ``cat'' and ``dog'' , respectively. 
It can be observed that the content generated in the local region aligns with the manually specified objects.
However, this naive method leads to the forced generation of objects that should not be in local region, affecting the model's ability to align with the control condition. The results for both cat and dog generation are bad.
The right image shows the superiority of Control Concept Matching. It adaptively matches the concept during denoising process. 
Figure \ref{fig:beta}(b) shows the impact of different \(\beta\) values on image quality.
Our focus is on stabilizing the selection of \(\text{C}^{t}_\text{control}\). 
A \(\beta\)  between 0.8 to 0.9 yields good results.

\section{Conclusions}
In this paper, we introduce local control as a new paradigm for image controllable generation, focusing on user-defined control of specific image regions. To mitigate local control dominance, we propose a training-free method that integrates smoothly with existing control models. This approach optimizes noised latents, enhancing object generation in non-local regions. We demonstrate a variety of results under the local control condition, successfully generating high-quality images faithful to the text prompts.

\bibliography{aaai25}

\begin{thebibliography}{40}
\providecommand{\natexlab}[1]{#1}

\bibitem[{Avrahami, Lischinski, and Fried(2022)}]{avrahami2022blended}
Avrahami, O.; Lischinski, D.; and Fried, O. 2022.
\newblock Blended diffusion for text-driven editing of natural images.
\newblock In \emph{Proceedings of the IEEE/CVF Conference on Computer Vision and Pattern Recognition}, 18208--18218.

\bibitem[{Cao et~al.(2023)Cao, Wang, Qi, Shan, Qie, and Zheng}]{masa}
Cao, M.; Wang, X.; Qi, Z.; Shan, Y.; Qie, X.; and Zheng, Y. 2023.
\newblock Masactrl: Tuning-free mutual self-attention control for consistent image synthesis and editing.
\newblock In \emph{Proceedings of the IEEE/CVF International Conference on Computer Vision}, 22560--22570.

\bibitem[{Chefer et~al.(2023)Chefer, Alaluf, Vinker, Wolf, and Cohen-Or}]{chefer2023attend}
Chefer, H.; Alaluf, Y.; Vinker, Y.; Wolf, L.; and Cohen-Or, D. 2023.
\newblock Attend-and-excite: Attention-based semantic guidance for text-to-image diffusion models.
\newblock \emph{ACM Transactions on Graphics (TOG)}, 42(4): 1--10.

\bibitem[{Dhariwal and Nichol(2021)}]{dhariwal2021diffusion}
Dhariwal, P.; and Nichol, A. 2021.
\newblock Diffusion models beat gans on image synthesis.
\newblock \emph{Advances in neural information processing systems}, 34: 8780--8794.

\bibitem[{Epstein et~al.(2023)Epstein, Jabri, Poole, Efros, and Holynski}]{epstein2023diffusion}
Epstein, D.; Jabri, A.; Poole, B.; Efros, A.; and Holynski, A. 2023.
\newblock Diffusion self-guidance for controllable image generation.
\newblock \emph{Advances in Neural Information Processing Systems}, 36: 16222--16239.

\bibitem[{Feng et~al.(2022)Feng, He, Fu, Jampani, Akula, Narayana, Basu, Wang, and Wang}]{feng2022training}
Feng, W.; He, X.; Fu, T.-J.; Jampani, V.; Akula, A.; Narayana, P.; Basu, S.; Wang, X.~E.; and Wang, W.~Y. 2022.
\newblock Training-free structured diffusion guidance for compositional text-to-image synthesis.
\newblock \emph{arXiv preprint arXiv:2212.05032}.

\bibitem[{Hertz et~al.(2022)Hertz, Mokady, Tenenbaum, Aberman, Pritch, and Cohen-Or}]{hertz2022prompt}
Hertz, A.; Mokady, R.; Tenenbaum, J.; Aberman, K.; Pritch, Y.; and Cohen-Or, D. 2022.
\newblock Prompt-to-Prompt Image Editing with Cross Attention Control.
\newblock \emph{arXiv preprint arXiv:2208.01626}.

\bibitem[{Ho, Jain, and Abbeel(2020)}]{ho2020denoising}
Ho, J.; Jain, A.; and Abbeel, P. 2020.
\newblock Denoising diffusion probabilistic models.
\newblock \emph{Advances in neural information processing systems}, 33: 6840--6851.

\bibitem[{Jain et~al.(2023)Jain, Li, Chiu, Hassani, Orlov, and Shi}]{oneformer}
Jain, J.; Li, J.; Chiu, M.~T.; Hassani, A.; Orlov, N.; and Shi, H. 2023.
\newblock Oneformer: One transformer to rule universal image segmentation.
\newblock In \emph{Proceedings of the IEEE/CVF Conference on Computer Vision and Pattern Recognition}, 2989--2998.

\bibitem[{Kirillov et~al.(2023)Kirillov, Mintun, Ravi, Mao, Rolland, Gustafson, Xiao, Whitehead, Berg, Lo et~al.}]{sam}
Kirillov, A.; Mintun, E.; Ravi, N.; Mao, H.; Rolland, C.; Gustafson, L.; Xiao, T.; Whitehead, S.; Berg, A.~C.; Lo, W.-Y.; et~al. 2023.
\newblock Segment anything.
\newblock In \emph{Proceedings of the IEEE/CVF International Conference on Computer Vision}, 4015--4026.

\bibitem[{Li et~al.(2023{\natexlab{a}})Li, Li, Savarese, and Hoi}]{li2023blip}
Li, J.; Li, D.; Savarese, S.; and Hoi, S. 2023{\natexlab{a}}.
\newblock Blip-2: Bootstrapping language-image pre-training with frozen image encoders and large language models.
\newblock \emph{arXiv preprint arXiv:2301.12597}.

\bibitem[{Li et~al.(2023{\natexlab{b}})Li, Liu, Wu, Mu, Yang, Gao, Li, and Lee}]{li2023gligen}
Li, Y.; Liu, H.; Wu, Q.; Mu, F.; Yang, J.; Gao, J.; Li, C.; and Lee, Y.~J. 2023{\natexlab{b}}.
\newblock Gligen: Open-set grounded text-to-image generation.
\newblock In \emph{Proceedings of the IEEE/CVF Conference on Computer Vision and Pattern Recognition}, 22511--22521.

\bibitem[{Lin et~al.(2014)Lin, Maire, Belongie, Hays, Perona, Ramanan, Doll{\'a}r, and Zitnick}]{coco}
Lin, T.-Y.; Maire, M.; Belongie, S.; Hays, J.; Perona, P.; Ramanan, D.; Doll{\'a}r, P.; and Zitnick, C.~L. 2014.
\newblock Microsoft coco: Common objects in context.
\newblock In \emph{Computer Vision--ECCV 2014: 13th European Conference, Zurich, Switzerland, September 6-12, 2014, Proceedings, Part V 13}, 740--755. Springer.

\bibitem[{Liu et~al.(2023)Liu, Zeng, Ren, Li, Zhang, Yang, Li, Yang, Su, Zhu et~al.}]{dino}
Liu, S.; Zeng, Z.; Ren, T.; Li, F.; Zhang, H.; Yang, J.; Li, C.; Yang, J.; Su, H.; Zhu, J.; et~al. 2023.
\newblock Grounding dino: Marrying dino with grounded pre-training for open-set object detection.
\newblock \emph{arXiv preprint arXiv:2303.05499}.

\bibitem[{Lu, Liu, and Kong(2023)}]{lu2023tf}
Lu, S.; Liu, Y.; and Kong, A. W.-K. 2023.
\newblock TF-ICON: Diffusion-Based Training-Free Cross-Domain Image Composition.
\newblock In \emph{Proceedings of the IEEE/CVF International Conference on Computer Vision}, 2294--2305.

\bibitem[{Lugmayr et~al.(2022)Lugmayr, Danelljan, Romero, Yu, Timofte, and Van~Gool}]{lugmayr2022repaint}
Lugmayr, A.; Danelljan, M.; Romero, A.; Yu, F.; Timofte, R.; and Van~Gool, L. 2022.
\newblock Repaint: Inpainting using denoising diffusion probabilistic models.
\newblock In \emph{Proceedings of the IEEE/CVF Conference on Computer Vision and Pattern Recognition}, 11461--11471.

\bibitem[{Mo et~al.(2024)Mo, Mu, Lin, Liu, Guan, Li, and Zhou}]{freecontrol}
Mo, S.; Mu, F.; Lin, K.~H.; Liu, Y.; Guan, B.; Li, Y.; and Zhou, B. 2024.
\newblock Freecontrol: Training-free spatial control of any text-to-image diffusion model with any condition.
\newblock In \emph{Proceedings of the IEEE/CVF Conference on Computer Vision and Pattern Recognition}, 7465--7475.

\bibitem[{Mou et~al.(2023)Mou, Wang, Xie, Zhang, Qi, Shan, and Qie}]{mou2023t2i}
Mou, C.; Wang, X.; Xie, L.; Zhang, J.; Qi, Z.; Shan, Y.; and Qie, X. 2023.
\newblock T2i-adapter: Learning adapters to dig out more controllable ability for text-to-image diffusion models.
\newblock \emph{arXiv preprint arXiv:2302.08453}.

\bibitem[{Nichol et~al.(2021)Nichol, Dhariwal, Ramesh, Shyam, Mishkin, McGrew, Sutskever, and Chen}]{nichol2021glide}
Nichol, A.; Dhariwal, P.; Ramesh, A.; Shyam, P.; Mishkin, P.; McGrew, B.; Sutskever, I.; and Chen, M. 2021.
\newblock Glide: Towards photorealistic image generation and editing with text-guided diffusion models.
\newblock \emph{arXiv preprint arXiv:2112.10741}.

\bibitem[{Patashnik et~al.(2023)Patashnik, Garibi, Azuri, Averbuch-Elor, and Cohen-Or}]{patashnik2023localizing}
Patashnik, O.; Garibi, D.; Azuri, I.; Averbuch-Elor, H.; and Cohen-Or, D. 2023.
\newblock Localizing object-level shape variations with text-to-image diffusion models.
\newblock In \emph{Proceedings of the IEEE/CVF International Conference on Computer Vision}, 23051--23061.

\bibitem[{Radford et~al.(2021{\natexlab{a}})Radford, Kim, Hallacy, Ramesh, Goh, Agarwal, Sastry, Askell, Mishkin, Clark et~al.}]{radford2021learning}
Radford, A.; Kim, J.~W.; Hallacy, C.; Ramesh, A.; Goh, G.; Agarwal, S.; Sastry, G.; Askell, A.; Mishkin, P.; Clark, J.; et~al. 2021{\natexlab{a}}.
\newblock Learning transferable visual models from natural language supervision.
\newblock In \emph{International conference on machine learning}, 8748--8763. PMLR.

\bibitem[{Radford et~al.(2021{\natexlab{b}})Radford, Kim, Hallacy, Ramesh, Goh, Agarwal, Sastry, Askell, Mishkin, Clark et~al.}]{clip}
Radford, A.; Kim, J.~W.; Hallacy, C.; Ramesh, A.; Goh, G.; Agarwal, S.; Sastry, G.; Askell, A.; Mishkin, P.; Clark, J.; et~al. 2021{\natexlab{b}}.
\newblock Learning transferable visual models from natural language supervision.
\newblock In \emph{ICML}, 8748--8763.

\bibitem[{Radford et~al.(2021{\natexlab{c}})Radford, Kim, Hallacy, Ramesh, Goh, Agarwal, Sastry, Askell, Mishkin, Clark et~al.}]{clips}
Radford, A.; Kim, J.~W.; Hallacy, C.; Ramesh, A.; Goh, G.; Agarwal, S.; Sastry, G.; Askell, A.; Mishkin, P.; Clark, J.; et~al. 2021{\natexlab{c}}.
\newblock Learning transferable visual models from natural language supervision.
\newblock In \emph{International conference on machine learning}, 8748--8763. PMLR.

\bibitem[{Raffel et~al.(2020)Raffel, Shazeer, Roberts, Lee, Narang, Matena, Zhou, Li, and Liu}]{raffel2020exploring}
Raffel, C.; Shazeer, N.; Roberts, A.; Lee, K.; Narang, S.; Matena, M.; Zhou, Y.; Li, W.; and Liu, P.~J. 2020.
\newblock Exploring the limits of transfer learning with a unified text-to-text transformer.
\newblock \emph{The Journal of Machine Learning Research}, 21(1): 5485--5551.

\bibitem[{Ramesh et~al.(2021)Ramesh, Pavlov, Goh, Gray, Voss, Radford, Chen, and Sutskever}]{ramesh2021zero}
Ramesh, A.; Pavlov, M.; Goh, G.; Gray, S.; Voss, C.; Radford, A.; Chen, M.; and Sutskever, I. 2021.
\newblock Zero-shot text-to-image generation.
\newblock In \emph{International Conference on Machine Learning}, 8821--8831. PMLR.

\bibitem[{Rombach et~al.(2022)Rombach, Blattmann, Lorenz, Esser, and Ommer}]{rombach2022high}
Rombach, R.; Blattmann, A.; Lorenz, D.; Esser, P.; and Ommer, B. 2022.
\newblock High-resolution image synthesis with latent diffusion models.
\newblock In \emph{Proceedings of the IEEE/CVF conference on computer vision and pattern recognition}, 10684--10695.

\bibitem[{Saharia et~al.(2022)Saharia, Chan, Saxena, Li, Whang, Denton, Ghasemipour, Gontijo~Lopes, Karagol~Ayan, Salimans et~al.}]{saharia2022photorealistic}
Saharia, C.; Chan, W.; Saxena, S.; Li, L.; Whang, J.; Denton, E.~L.; Ghasemipour, K.; Gontijo~Lopes, R.; Karagol~Ayan, B.; Salimans, T.; et~al. 2022.
\newblock Photorealistic text-to-image diffusion models with deep language understanding.
\newblock \emph{Advances in Neural Information Processing Systems}, 35: 36479--36494.

\bibitem[{Schuhmann et~al.(2022)Schuhmann, Beaumont, Vencu, Gordon, Wightman, Cherti, Coombes, Katta, Mullis, Wortsman et~al.}]{schuhmann2022laion}
Schuhmann, C.; Beaumont, R.; Vencu, R.; Gordon, C.; Wightman, R.; Cherti, M.; Coombes, T.; Katta, A.; Mullis, C.; Wortsman, M.; et~al. 2022.
\newblock Laion-5b: An open large-scale dataset for training next generation image-text models.
\newblock \emph{Advances in Neural Information Processing Systems}, 35: 25278--25294.

\bibitem[{Seitzer(2020)}]{fid}
Seitzer, M. 2020.
\newblock {pytorch-fid: FID Score for PyTorch}.
\newblock \url{https://github.com/mseitzer/pytorch-fid}.
\newblock Version 0.3.0.

\bibitem[{Tumanyan et~al.(2023)Tumanyan, Geyer, Bagon, and Dekel}]{Plug-and-Play_2023_CVPR}
Tumanyan, N.; Geyer, M.; Bagon, S.; and Dekel, T. 2023.
\newblock Plug-and-Play Diffusion Features for Text-Driven Image-to-Image Translation.
\newblock In \emph{Proceedings of the IEEE/CVF Conference on Computer Vision and Pattern Recognition (CVPR)}, 1921--1930.

\bibitem[{Tunanyan et~al.(2023)Tunanyan, Xu, Navasardyan, Wang, and Shi}]{tunanyan2023multi}
Tunanyan, H.; Xu, D.; Navasardyan, S.; Wang, Z.; and Shi, H. 2023.
\newblock Multi-Concept T2I-Zero: Tweaking Only The Text Embeddings and Nothing Else.
\newblock \emph{arXiv preprint arXiv:2310.07419}.

\bibitem[{Voynov, Aberman, and Cohen-Or(2023)}]{voynov2023sketch}
Voynov, A.; Aberman, K.; and Cohen-Or, D. 2023.
\newblock Sketch-guided text-to-image diffusion models.
\newblock In \emph{ACM SIGGRAPH 2023 Conference Proceedings}, 1--11.

\bibitem[{Wang et~al.(2022)Wang, Zhang, Zhang, Ouyang, Chen, Chen, and Wen}]{wang2022pretraining}
Wang, T.; Zhang, T.; Zhang, B.; Ouyang, H.; Chen, D.; Chen, Q.; and Wen, F. 2022.
\newblock Pretraining is all you need for image-to-image translation.
\newblock \emph{arXiv preprint arXiv:2205.12952}.

\bibitem[{Xie et~al.(2023)Xie, Li, Huang, Liu, Zhang, Zheng, and Shou}]{xie2023boxdiff}
Xie, J.; Li, Y.; Huang, Y.; Liu, H.; Zhang, W.; Zheng, Y.; and Shou, M.~Z. 2023.
\newblock Boxdiff: Text-to-image synthesis with training-free box-constrained diffusion.
\newblock In \emph{Proceedings of the IEEE/CVF International Conference on Computer Vision}, 7452--7461.

\bibitem[{Xu et~al.(2018)Xu, Zhang, Huang, Zhang, Gan, Huang, and He}]{xu2018attngan}
Xu, T.; Zhang, P.; Huang, Q.; Zhang, H.; Gan, Z.; Huang, X.; and He, X. 2018.
\newblock Attngan: Fine-grained text to image generation with attentional generative adversarial networks.
\newblock In \emph{Proceedings of the IEEE conference on computer vision and pattern recognition}, 1316--1324.

\bibitem[{Yang et~al.(2023)Yang, Gu, Zhang, Zhang, Chen, Sun, Chen, and Wen}]{yang2023paint}
Yang, B.; Gu, S.; Zhang, B.; Zhang, T.; Chen, X.; Sun, X.; Chen, D.; and Wen, F. 2023.
\newblock Paint by example: Exemplar-based image editing with diffusion models.
\newblock In \emph{Proceedings of the IEEE/CVF Conference on Computer Vision and Pattern Recognition}, 18381--18391.

\bibitem[{Zhang et~al.(2017)Zhang, Xu, Li, Zhang, Wang, Huang, and Metaxas}]{zhang2017stackgan}
Zhang, H.; Xu, T.; Li, H.; Zhang, S.; Wang, X.; Huang, X.; and Metaxas, D.~N. 2017.
\newblock Stackgan: Text to photo-realistic image synthesis with stacked generative adversarial networks.
\newblock In \emph{Proceedings of the IEEE international conference on computer vision}, 5907--5915.

\bibitem[{Zhang, Rao, and Agrawala(2023)}]{zhang2023adding}
Zhang, L.; Rao, A.; and Agrawala, M. 2023.
\newblock Adding conditional control to text-to-image diffusion models.
\newblock In \emph{Proceedings of the IEEE/CVF International Conference on Computer Vision}, 3836--3847.

\bibitem[{Zhang et~al.(2018)Zhang, Isola, Efros, Shechtman, and Wang}]{lpips}
Zhang, R.; Isola, P.; Efros, A.~A.; Shechtman, E.; and Wang, O. 2018.
\newblock The unreasonable effectiveness of deep features as a perceptual metric.
\newblock In \emph{Proceedings of the IEEE conference on computer vision and pattern recognition}, 586--595.

\bibitem[{Zhao et~al.(2024)Zhao, Chen, Chen, Bao, Hao, Yuan, and Wong}]{uni}
Zhao, S.; Chen, D.; Chen, Y.-C.; Bao, J.; Hao, S.; Yuan, L.; and Wong, K.-Y.~K. 2024.
\newblock Uni-controlnet: All-in-one control to text-to-image diffusion models.
\newblock \emph{Advances in Neural Information Processing Systems}, 36.

\end{thebibliography}
\end{document}